\documentclass{article}

\usepackage[preprint,nonatbib]{neurips_2022}

\usepackage[utf8]{inputenc} 
\usepackage[T1]{fontenc}    
\usepackage{hyperref}       
\usepackage{url}            
\usepackage{booktabs}       
\usepackage{amsfonts}       
\usepackage{nicefrac}       
\usepackage{microtype}      
\usepackage{xcolor}         

\hypersetup{
    colorlinks=true,
    linkcolor=red,
    citecolor=cyan,
    filecolor=magenta,      
    urlcolor=cyan,
    }
\usepackage{subcaption}
\usepackage{graphicx}
\usepackage{multirow}
\usepackage{pifont}
\newcommand{\cmark}{\ding{51}}%
\newcommand{\xmark}{\ding{55}}%

\newcommand{\ie}{\textit{i}.\textit{e}.}
\newcommand{\eg}{\textit{e}.\textit{g}.}
\newcommand{\vs}{\textit{v}.\textit{s}.}
\newcommand{\etc}{\textit{etc}.}
\definecolor{00red}{RGB}{236,35,35}
\newcommand{\whred}[1]{\textcolor{00red}{#1}}
\definecolor{00blue}{RGB}{50,149,237}

\definecolor{00green}{RGB}{82,208,83}

\definecolor{02pink}{RGB}{240,178,188}

\newcommand{\cellblue}[1]{\cellcolor{00blue!30}{#1}}

\definecolor{demphcolor1}{gray}{.6}

\definecolor{demphcolor}{RGB}{144,144,144}

\usepackage{colortbl}
\definecolor{mygray}{gray}{0.95}
\newlength\savewidth
\usepackage{arydshln} 
\usepackage[ruled]{algorithm2e}
\usepackage{listings}
\usepackage{fontawesome}
\usepackage{makecell}
\usepackage{bm}
\usepackage{amsmath}
\usepackage{cleveref}

\title{\emph{FreeVA}: Offline MLLM as Training-Free Video Assistant}

\author{
Wenhao Wu \\
The University of Sydney \\
{\tt\small \url{https://github.com/whwu95/FreeVA}}
}

\begin{document}
\maketitle


\begin{abstract}
This paper undertakes an empirical study to revisit the latest advancements in Multimodal Large Language Models (MLLMs): Video Assistant. This study, namely \textbf{FreeVA}, aims to extend existing image-based MLLM to the video domain in a training-free manner.
The study provides an essential, yet must-know baseline, and reveals several surprising findings: 
1) FreeVA, leveraging only offline image-based MLLM without additional training, excels in zero-shot video question-answering (\eg, MSVD-QA, ActivityNet-QA, and MSRVTT-QA), even surpassing state-of-the-art methods that involve video instruction tuning. 
2) While mainstream video-based MLLMs typically initialize with an image-based MLLM (\eg, LLaVA) and then fine-tune using video instruction tuning, the study indicates that utilizing the widely adopted VideoInstruct-100K for video instruction tuning doesn't actually lead to better performance compared to not training at all.
3) The commonly used evaluation metrics in existing works are significantly influenced by changes in the GPT API version over time. If ignored, this could affect the fairness and uniformity of comparisons between different methods and impact the analysis and judgment of researchers in the field.
The advancement of MLLMs is currently thriving, drawing numerous researchers into the field.
We aim for this work to serve as a plug-and-play, simple yet effective baseline, encouraging the direct evaluation of existing MLLMs in video domain while also standardizing the field of video conversational models to a certain extent. Also, we encourage researchers to reconsider: \emph{Have current video MLLM methods truly acquired knowledge beyond image MLLM?}
Code is available at \url{https://github.com/whwu95/FreeVA}.
\end{abstract}

\section{Introduction}

In late 2022, the emergence of ChatGPT~\cite{ChatGPT} sparked a new wave of artificial intelligence revolution, elevating the discourse around large language models (LLMs) to new heights. In 2023, OpenAI further advanced the field by introducing GPT-4V~\cite{GPT4}, a multimodal large language model (MLLM) endowed with image understanding capabilities. This development garnered widespread attention in MLLMs, both in academia and industry, making it a burgeoning research hotspot.

In pursuit of keeping pace with GPT-4V, researchers collectively introduced a plethora of GPT-4V-like MLLMs (\eg, LLaVA~\cite{llava}, Mini-GPT4~\cite{minigpt4}, InstructBLIP~\cite{instructblip}, \etc). These models demonstrated promising performance across a range of multi-modal tasks, such as image captioning and visual reasoning, gradually forming a unified model design and training paradigm. Structurally, these models typically comprise a pre-trained vision encoder (\eg, CLIP~\cite{CLIP,evaclip} pre-trained ViT~\cite{CLIP}) for encoding image into visual tokens, a large language model (\eg, OPT~\cite{zhang2022opt}, Vicuna~\cite{vicuna}, \etc) enriched with extensive knowledge, and a vision-language connector (\eg, Q-former~\cite{blip2}, MLP projection~\cite{llava}) enabling the language model to understand visual tokens. In terms of training, the process typically comprises two stages: first, pre-training the connector for visual-text alignment, followed by the second stage of visual instruction tuning to imbue the model with instruction understanding capabilities.
In contrast to these image MLLMs, the development of video MLLMs has lagged behind. Typically, these video MLLMs~\cite{videochatgpt,li2023llamavid,luo2023valley,wang2023vaquita} initialize with an image MLLM and then fine-tuning using video-text datasets. VideoChatGPT~\cite{videochatgpt} stands as a pioneering example, introducing a video instruction tuning dataset, VideoInstruct100K~\cite{videochatgpt}, to transfer well-trained LLaVA~\cite{llava} to video domain. It also introduced the first quantitative evaluation benchmark for video-based conversational models, moving away from qualitative case studies. Subsequent works primarily employ VideoInstruct100K for instruction tuning and evaluate their models using this benchmark.

Currently, research in MLLMs primarily focuses on images, with larger publicly available image-text datasets being continually introduced and more powerful models emerging. Therefore, we wondered whether advanced MLLMs, without additional video tuning, already possess the knowledge necessary for video understanding. Based on this premise, this paper conducts an empirical study called \textbf{\whred{FreeVA}}, aimed at exploring the use of existing image-based MLLMs directly as training-\whred{free} \whred{v}ideo \whred{a}ssistants.

To achieve this, we proposed a straightforward approach: each video frame undergoes a processing similar to the inference process of the image MLLM. the vision encoder and vision-language connector generate visual tokens that are ``understandable'' by the language model. We simply added a parameter-free temporal aggregation to aggregate these tokens temporally, which are subsequently directly fed to the language model.
The findings of the study surprised us in several ways: 
\begin{enumerate}
    \item Offline image MLLMs, when combined with proper temporal aggregation, achieved state-of-the-art performance in zero-shot video question-answering tasks (\eg, MSVD-QA~\cite{msvd}, ActivityNet-QA~\cite{activitynet}, and MSRVTT-QA~\cite{msrvtt}). This performance even surpassed that of previous methods trained using video instruction tuning.
    \item While mainstream video MLLMs~\cite{videochatgpt,luo2023valley,wang2023vaquita} typically initialize with an well-trained image MLLM (\eg, LLaVA~\cite{llava}) and then fine-tune using video instruction tuning, we discovered a counterintuitive and previously overlooked phenomenon: fine-tuning LLaVA with the existing VideoInstruct-100K dataset for video instruction tuning resulted in worse performance on video question-answering tasks compared to the original LLaVA (see \Cref{tab:off_vs_on} and \Cref{tab:videochatgpt}). This discovery suggests that the community needs to reassess the effectiveness of the widely used VideoInstruct-100K dataset and whether video MLLMs have indeed acquired more knowledge than initialized image MLLMs.
    \item We discovered a significant yet overlooked issue: the commonly used evaluation metrics for zero-shot video question-answering are heavily influenced by changes in the GPT-3.5 API version. Given that the default version of the API has been changing over time, this could greatly affect the fairness of comparisons between methods across different time periods.
\end{enumerate}

We understand that these findings may challenge the prevailing beliefs in the field and potentially impact the experimental conclusions or performance comparisons of some published or forthcoming works. However, we think it's crucial to shed light on these matters for the sake of the community's future development. Based on this study, it appears that current video MLLMs, despite utilizing video-text datasets for tuning, may not have surpassed image MLLMs in practice. We hope that FreeVA can serve as a plug-and-play tool to encourage more existing image MLLMs to be directly evaluated on video tasks. We also encourage
researchers to pause and rethink: \emph{Have current video MLLM methods truly acquired knowledge beyond image MLLMs?}

\section{Related Works}

\textbf{Large Language Models.} The emergence of LLMs (\eg, GPT~\cite{gpt,GPT3}, LLaMA~\cite{llama,llama2}, OPT~\cite{zhang2022opt}, \etc) has transformed Natural Language Processing (NLP). These models, based on the Transformer architecture, excel in tasks such as language generation and in-context learning. Their ability to understand complex prompts in a zero-shot manner showcases their adaptability and generalization prowess. To unlock their full potential, instruction tuning has become crucial, focusing on aligning models with user intentions and enhancing output quality. Models like InstructGPT~\cite{instructgpt} and ChatGPT~\cite{ChatGPT} exemplify this strategy, leading to significant performance enhancements. Open-source models like Alpaca~\cite{alpaca} and Vicuna~\cite{vicuna}, leveraging the LLaMA framework, have demonstrated notable improvements through instruction tuning. Ongoing research also explores utilizing LLMs' reasoning abilities for visual applications~\cite{wu2023gpt4vis,dai2024gpt4ego,wu2023cap4video}, broadening their utility across various domains.

\textbf{Vision-Language Models.}
Vision-language cross-modal learning, pioneered by CLIP~\cite{CLIP}, has been a hot topic in recent years. CLIP achieves this through 400M vision-text contrastive training, simultaneously training vision and text encoders to bridge the gap between visual and textual modalities. This has empowered visual models with potent zero-shot capabilities, spurring subsequent research in image-text alignment~\cite{yuan2021florence,ALIGN} and video-text alignment~\cite{fanguatvr,wu2023cap4video,text4vis,BIKE}. With the rise of LLMs, there's a growing emphasis on further integrating LLM knowledge into visual models, known as Multimodal Large Language Models (MLLMs). 
During this period, CLIP's pre-trained vision encoder remains mainstream and is often frozen, while research focuses on connecting the vision encoder with LLMs (\eg, BLIP2~\cite{blip2}, Flamingo~\cite{flamingo}). Additionally, there's a growing focus on visual instruction tuning~\cite{minigpt4,llava,instructblip,ye2023mplug} to enable models to answer based on instructions, similar to GPT-4V. 
For instance, MiniGPT-4~\cite{minigpt4} constructs instruction pairs from BLIP-2, facilitating image-based conversations by integrating with Vicuna~\cite{vicuna}. Similarly, LLaVA~\cite{llava,llava1.5} collects larger datasets to train a simple linear projector to align the image and text space of LLM, demonstrating visual reasoning capabilities.
Other methods like InstructBLIP~\cite{instructblip} and mPLUG-Owl~\cite{ye2023mplug} also benefit from vision instruction tuning.
Naturally, VideoChat~\cite{videochat} and Video-LLaMA~\cite{videollama} aim to facilitate video-based conversation, however, both approaches lack quantitative results. Furthermore, Video-ChatGPT~\cite{videochatgpt} introduces a dataset of 100k video instructions for fine-tuning LLaVA to the video domain, along with the first quantitative evaluation benchmark for video-based conversational models. Subsequent research~\cite{li2023llamavid,luo2023valley,wang2023vaquita,Video-LLaVA} largely follows this approach, conducting video instruction tuning and evaluating their models using this benchmark. Additionally, MovieChat~\cite{moviechat} designs a complex short-term and long-term memory mechanism to extend Video-LLaMA~\cite{videollama} for long video understanding and introduces the MovieChat-1K dataset for training and evaluating long video question-answering tasks.
In contrast to these tuning approaches, our work primarily focuses on exploring training-free video-based conversational models. Research in this domain is relatively limited. This paper aims to provide a simple and effective plug-and-play strategy to extend image-based MLLMs to general video scenarios. It quantitatively compares with tuning methods and offers previous overlooked experience.

\section{Method}

\subsection{Preliminaries: MLLM Architecture Overview}
\begin{figure}[!t]
  \centering
  \begin{subfigure}[b]{\textwidth}
    \centering
    \includegraphics[width=0.85\textwidth]{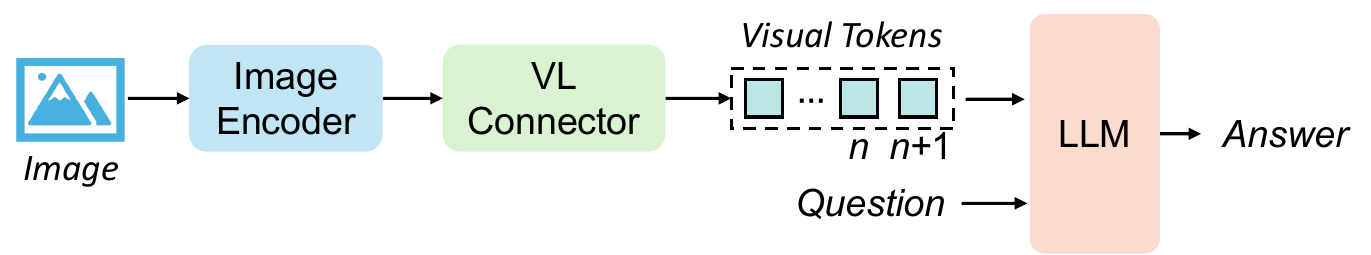}
    \caption{Inference workflow of image MLLMs (\eg, BLIP2~\cite{blip2}, LLaVA~\cite{llava}). An input image is first processed by the \emph{Image Encoder} (\eg, ViT-L~\cite{CLIP}) to extract visual features, which are then converted into language embeddings by the \emph{Vision-Language (VL) Connector} (\eg, Q-former~\cite{blip2}, projection~\cite{llava}). Finally, the \emph{LLM} (\eg, Vicuna~\cite{vicuna}) interprets these visual tokens to answer questions. Here, $n$ represents the index of patch tokens.}
    \vspace{2mm}
    \label{fig:image_llm}
  \end{subfigure}
  \begin{subfigure}[b]{\textwidth}
    \centering
    \includegraphics[width=0.89\textwidth]{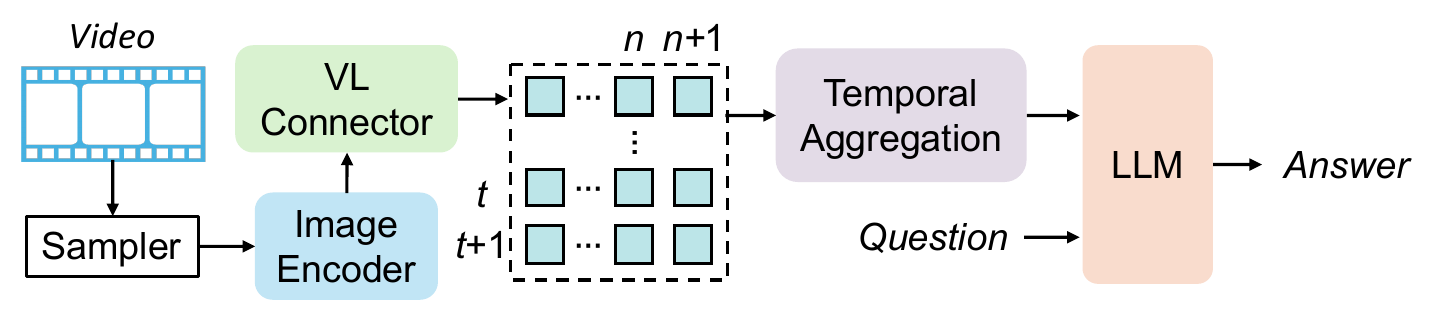}
    \caption{FreeVA: A training-free pipeline for video question answering using existing image MLLMs. Here, $t$ indicates the index of the sampled frames. \emph{Too simple? That's enough!}}
    \label{fig:video_llm}
  \end{subfigure}
   \caption{An illustration of (a) an overview of the image MLLM inference process and (b) our proposed FreeVA for zero-shot video inference using existing image MLLMs.}
  \label{fig:overview}
\end{figure}

To adapt trained MLLMs for the video domain, let's briefly review the current structure of MLLMs. 
As depicted in \Cref{fig:image_llm}, mainstream MLLMs are typically comprised of three key components: an image encoder (\eg, CLIP's ViT-L~\cite{CLIP} or EVA-CLIP's ViT-G~\cite{evaclip}), a Vision-Language (VL) Connector (\eg, Q-former~\cite{blip2} or MLP projection~\cite{llava}), and a Large Language Model (LLM) (\eg, OPT~\cite{zhang2022opt}, Vicuna~\cite{vicuna}, \etc). 
During the inference phase, an input image is initially processed by the image backbone to extract its visual features. The vision-language connector then projects these visual features into language embeddings that align with the LLM, thus enabling the LLM to understand visual tokens. Finally, the LLM uses these tokens to generate predictions in the form of text.

\subsection{FreeVA: Training-Free Video Assistant}
Compared to the development of image-based MLLM, the progress of video-based MLLM has been relatively slower. Its training paradigm largely follows that of image MLLMs, and both the training data for video-text pairs and video instruction tuning are significantly less abundant than in the image domain. Given the current shortage of high-quality data resources available for training video MLLM, and considering the continuous emergence and updates of excellent image MLLMs (\eg, LLaVA~\cite{llava}, InternVL~\cite{internvl-1.5}, DC~\cite{dc}), it prompts us to ponder: \emph{Can we leverage these well-trained, outstanding Image MLLMs directly for video modality without any training (\ie, in a zero-shot manner)?}

Figure~\ref{fig:video_llm} presents our FreeVA, which is remarkably simple and intuitive. 
Compared to the inference process of image MLLM, our FreeVA requires only minor adjustments. Specifically, we use a video sampler to capture multiple frames from the video as inputs and design a parameter-free temporal aggregation mechanism to aggregate visual tokens across frames.
Clearly, the key aspect of FreeVA is temporal aggregation. Formally, given visual tokens $X^\texttt{IN} \in {\mathbb{R}}^{T\times N \times D}$ from multiple frames, where $T$ represents the number of frames, $N$ is the number of patches per frame, and $D$ is the feature dimension, we need to aggregate $X^\texttt{IN} \in {\mathbb{R}}^{T\times N \times D} \rightarrow X^\texttt{OUT} \in {\mathbb{R}}^{K \times D}$ before feeding them to the LLM, where $K$ is the number of aggregated tokens. Here we discuss two intuitive types of aggregation as follows:

\textbf{Sparse Aggregation:} Global Average Pooling (GAP) serves as a straightforward aggregation method. In this paper, I loosely categorize such methods as Sparse Aggregation. In video understanding, mean pooling across different frames in the temporal dimension is a commonly used technique to obtain a global video representation. In this scenario (\ie, $X^\texttt{IN} \in {\mathbb{R}}^{T\times N \times D} \xrightarrow{\textbf{T-GAP}} X^\texttt{OUT} \in {\mathbb{R}}^{N \times D}$), the number of aggregated tokens $K$, equals $N$. For clarity in subsequent experimental \Cref{sec:ablation}, I refer to this method as \texttt{S1}. Similarly, we can compress features in the spatial dimension (\ie, (\ie, $X^\texttt{IN} \in {\mathbb{R}}^{T\times N \times D} \xrightarrow{\textbf{S-GAP}} X^\texttt{OUT} \in {\mathbb{R}}^{T \times D}$)), where $K$ equals $T$. I denote this method as \texttt{S2}. By combining both \texttt{S1} and \texttt{S2} methods (\ie, $X^\texttt{IN} \in {\mathbb{R}}^{T\times N \times D} \rightarrow X^\texttt{OUT} \in {\mathbb{R}}^{(T+N)\times D}$), we obtain $N+T$ aggregated tokens, referred to as \texttt{S3}. 
It's worth noting that \texttt{S3} was also used in the model training of VideoChatGPT~\cite{videochatgpt}. However, in this paper, we focus on the training-free scenario. Interestingly, empirical evidence (as shown in \Cref{tab:more_agg} in \Cref{sec:ablation}) demonstrates that \texttt{S3} performs worse than \texttt{S1}, and all sparse aggregations perform much worse than the dense aggregation introduced below.

\textbf{Dense Aggregation:} Unlike directly compressing the temporal or spatial dimensions through GAP to obtain sparse tokens, I also consider to retain more visual tokens. A straightforward approach \texttt{D1} (\ie, $X^\texttt{IN} \in {\mathbb{R}}^{T\times N \times D} \xrightarrow{\textbf{Concat}} X^\texttt{OUT} \in {\mathbb{R}}^{(TN) \times D}$), involves preserving all visual tokens from multiple frames, resulting in $K$ being equal to $T\times N$. Furthermore, considering the input token limit of the LLM, retaining all tokens from each frame would restrict the number of input frames. Therefore, I aim to compress the number of tokens per frame to include information from more frames. Specifically, in \texttt{D2} (\ie, $X^\texttt{IN} \in {\mathbb{R}}^{\alpha T\times N \times D} \xrightarrow{\textbf{Spatial Pool}} X \in {\mathbb{R}}^{\alpha T\times N/\alpha \times D} \xrightarrow{\textbf{Concat}} X^\texttt{OUT} \in {\mathbb{R}}^{(TN) \times D}$), I use max pooling with a stride of $\alpha$ to reduce the number of tokens per frame from $N$ to $N/\alpha$, while simultaneously scaling the sampled frames to maintain the same number of aggregated tokens.

After obtaining the aggregated tokens, we feed them directly into the LLM, which then generates responses based on user instructions.

\section{Experiments}
\subsection{Experimental Setup}
\noindent\textbf{Implementation Details.}
All experiments can be completed using a single 40G A100 GPU.
In this paper, we utilize the existing image-based MLLMs (\eg, LLaVA-1.5~\cite{llava1.5}, InstructBLIP~\cite{instructblip}, InternVL~\cite{internvl-1.5}, Dense Connector~\cite{dc}) as examples to extend their capabilities to the video domain in a training-free manner, enabling direct zero-shot video understanding. We believe that the performance can be further improved as the image MLLMs become more powerful. 
We uniformly sample the entire video to acquire $T$ frames as input, this can be increased as the token limit of the LLM allows.
 
\noindent\textbf{Datasets.} 
Our method requires no training. We evaluate zero-shot performance on open-ended video question-answering benchmarks, \eg, MSVD-QA~\cite{msvd}, ActivityNet-QA~\cite{activitynet}, MSRVTT-QA~\cite{msrvtt}, and the newly proposed generative performance benchmark~\cite{videochatgpt}, include metrics such as Correctness of Information (\textbf{CI}), Detail Orientation (\textbf{DO}), Contextual Understanding (\textbf{CU}), Temporal Understanding (\textbf{TU}), and Consistency (\textbf{CO}). The evaluation process for video understanding aligns with Video-ChatGPT~\cite{videochatgpt}. Accuracy and score are reported, with assessments conducted using GPT-3.5 assistant.
The next section will reveal a previously overlooked aspect: the impact of the GPT-3.5 version on evaluation.

\subsection{Empirical Observations on Basic Factors}
\label{sec:ablation}
In this section, we study how the basic factors influence video question-answering performance.

\textbf{Sparse $\vs$ Dense Temporal Aggregation.}
Table~\ref{tab:num_frames} presents the results of two types of temporal aggregation with varying numbers of video frames. 1) \emph{Sparse Temporal Aggregation:} As shown in Table~\ref{tab:sparse}, increasing the number of video frames does not consistently enhance performance. Surprisingly, using 100 frames does not yield better results compared to using just one frame, despite 100 frames being a common setting in previous works (\eg, VideoChatGPT~\cite{videochatgpt}, Valley~\cite{luo2023valley}). Notably, on ActivityNet-QA, good performance can be achieved with just one frame, and adding more frames significantly degrades performance. On MSVD-QA, using 4 frames leads to a slight improvement, but further increasing the number of frames results in performance degradation. This may be because the visual tokens produced by the visual encoder of the offline image MLLM can already be "understood" by the LLM. Averaging across frames in the temporal dimension may disrupt the features of each frame's patch tokens, and with more frames involved, this disruption increases, thereby diminishing the LLM's understanding of visual tokens.
2) \emph{Dense Temporal Aggregation:} As shown in Table~\ref{tab:dense}, increasing the number of video frames significantly improves performance, surpassing the performance achieved with sparse temporal aggregation by a large margin. However, increasing the number of frames to 8 causes performance to collapse due to exceeding the LLM's token limit. This suggests that increasing the number of visual tokens within the LLM's token limit is a viable way to enhance performance. Thus, using 4 frames yielded the best results on both datasets.

\begin{table}[t]
     \caption{Exploration of two intuitive temporal aggregation methods on MSVD-QA and ActivityNet-QA datasets. \texttt{GPT-3.5-Turbo-0613} API is employed for evaluation. Here, $T$ denotes the number of input frames, $N$ denotes the number of patch tokens per frame, and $D$ denotes the feature dimension.}
     \vspace{1mm}
    \begin{subtable}[t]{0.47\textwidth}
        \centering
        \scalebox{0.92}{
        \begin{tabular}{ccccc}
        \toprule
        \multirow{2}{*}{\#Frame}  & \multicolumn{2}{c}{\bf MSVD-QA} & \multicolumn{2}{c}{\bf ActivityNet-QA} \\
          &  Acc & Score &  Acc & Score  \\ \midrule
        1 & 68.9 & 3.72 & \cellblue{\textbf{51.5}} & \cellblue{\textbf{3.34}} \\
        4 & \cellblue{\textbf{69.6}} & \cellblue{\textbf{3.74}} & 51.0 & 3.35 \\
        8 & 68.8 & 3.72 & 49.2 & 3.30 \\
        100 & 67.3 & 3.67 & 46.2 & 3.22 \\
        \bottomrule
        \end{tabular}}
        \caption{\textbf{Sparse Temporal Aggregation}: \emph{T$\times$N$\times$D → Mean Pooling → N$\times$D.}}
       \label{tab:sparse}
    \end{subtable}
    \hfill
    \begin{subtable}[t]{0.47\textwidth}
        \centering
        \scalebox{0.92}{
        \begin{tabular}{ccccc}
        \toprule
        \multirow{2}{*}{\#Frame}  & \multicolumn{2}{c}{\bf MSVD-QA} & \multicolumn{2}{c}{\bf ActivityNet-QA} \\
          &  Acc & Score  &  Acc & Score \\ \midrule
        1 & 68.9 & 3.72 & 51.5 & 3.34 \\
        2 & 69.5 & 3.73 & 53.1 & 3.40 \\
        4 & \cellblue{\textbf{70.5}} & \cellblue{\textbf{3.77}} & \cellblue{\textbf{53.5}} & \cellblue{\textbf{3.42}} \\
        8 & 25.0 & 1.25 & 12.8 & 1.11 \\
        \bottomrule
        \end{tabular}}
        \caption{\textbf{Dense Temporal Aggregation}: \emph{T$\times$N$\times$D → Concatenation → TN$\times$D.}}
        \label{tab:dense}
     \end{subtable}
     \label{tab:num_frames}
\end{table}

\begin{table}[t]
    \begin{center}
    \caption{Impact of different temporal aggregation on MSVD-QA dataset evaluation. ``After Proj.'' indicates whether aggregation is performed after Vision-Language Connector (\ie, the projection in LLaVA). \texttt{GPT-3.5-Turbo-0613} API is employed for evaluation. Here, $T=4$, $N=576$.}
     \label{tab:more_agg}
    \vspace{1mm}
    \setlength{\tabcolsep}{2pt}
    \scalebox{0.9}{
    \begin{tabular}{cllcc}
    \toprule
    After Proj.&  & Aggregation & Acc & Score \\ \midrule 
    \xmark & S1: & \emph{T$\times$N$\times$D $\xrightarrow{\color{teal}\textbf{Temporal Mean Pool}}$  N$\times$D}  & 68.6 & 3.72 \\ \midrule
    \cmark & S1: &\emph{T$\times$N$\times$D $\xrightarrow{\color{teal}\textbf{Temporal Mean Pool}}$ N$\times$D}  & 69.6 & 3.74 \\
    \cmark & S2: &\emph{T$\times$N$\times$D $\xrightarrow{\color{teal}\textbf{Spatial Mean Pool}}$ T$\times$D}  & 47.4 & 3.64 \\
    \cmark & S3: &\emph{T$\times$N$\times$D $\xrightarrow{\color{teal}\textbf{Spatail+Temporal Mean Pool}}$ (N+T)$\times$D} & 68.7 & 3.72 \\  \midrule
    \cmark & D1: &\emph{T$\times$N$\times$D  $\xrightarrow{\color{teal}\textbf{Concat}}$ (TN)$\times$D}  & 70.5 & 3.77 \\ 
    \cmark  & D2:  &\emph{2T$\times$N$\times$D $\xrightarrow{\color{teal}\textbf{Spatial Max Pool (stride 2)}}$ 2T$\times$N/2$\times$D $\xrightarrow{\color{teal}\textbf{Concat}}$ (TN)$\times$D} &  \cellblue{\textbf{71.2}} & \cellblue{\textbf{3.81}} \\
    \bottomrule
    \end{tabular}}                
    \end{center}
\end{table}

\begin{table}[t]
    \begin{minipage}{0.48\textwidth}
    \begin{center}
    \caption{
Pre-trained LLaVA-1.5 \emph{vs.} LLaVA-1.5 fine-tuned with video instruction tuning (\ie, VideoInstruct-100K dataset~\cite{videochatgpt}). \faLock~indicates parameters are frozen. Setting: Sparse temporal aggregation with 4 frames.}
\label{tab:off_vs_on}   
    \vspace{1mm}
    \setlength{\tabcolsep}{4pt}
    \scalebox{0.9}{
    \begin{tabular}{lcccc}
    \toprule
    &  \multirow{2}{*}{Projection} & \multirow{2}{*}{LLM}  &  \multicolumn{2}{c}{\textbf{MSVD-QA}} \\
         &  &  & Acc & Score \\ \midrule
    Offline   & \faLock & \faLock & \cellblue{\textbf{69.6}} & \cellblue{\textbf{3.74}}  \\ \midrule
    \multirow{2}{*}{Online}  & \faUnlock & \faLock & 69.0 & 3.72 \\
      &    \faUnlock & \faUnlock & 57.1 & 3.36 \\ 
    \bottomrule
    \end{tabular}}        
    \end{center}     
    \end{minipage}
    \hspace{2mm}
    \begin{minipage}{0.48\textwidth}
    \begin{center}
    \setlength{\tabcolsep}{3pt}
      \caption{VideoChatGPT~\cite{videochatgpt} Initial \vs~Trained Parameters. It starts with LLaVA-v1 and updates only the projection. \texttt{GPT-3.5-Turbo-0613} API is employed for evaluation.}
      \label{tab:videochatgpt}
      \vspace{1mm}
      \scalebox{0.9}{
    \begin{tabular}{lccc}
    \toprule
    \multirow{2}{*}{VideoChatGPT} & \multirow{2}{*}{\#Frames} & \multicolumn{2}{c}{\bf MSVD-QA} \\
     &  &  Acc & Score  \\ \midrule
      Trained Param. & 100 & 67.1 & 3.68 \\
      Trained Param. & 4 & 67.8 & 3.69  \\ \midrule
      Initialized Param. & 100 & 68.1 & 3.70  \\ 
      Initialized Param. & 4 & \cellblue{\textbf{69.2}} & \cellblue{\textbf{3.72}}  \\   
      \bottomrule
    \end{tabular}}
    \end{center}
    \end{minipage}
\end{table}

\begin{table}[t]
    \begin{minipage}{0.38\textwidth}
    \begin{center}
    \caption{FreeVA with various sizes of LLMs. \texttt{GPT-3.5-Turbo-0613} API is employed for evaluation.}
    \label{tab:llm}
    \vspace{1mm}
    \setlength{\tabcolsep}{1pt}
    \scalebox{0.88}{
    \begin{tabular}{lccccc}
    \toprule
    \multirow{2}{*}{LLM } & \multirow{2}{*}{Agg.} & \multicolumn{2}{c}{\bf MSVD-QA} & \multicolumn{2}{c}{\bf ANet-QA} \\
     &  &  Acc & Score &  Acc & Score  \\ \midrule
      Vicuna-7B & D1 & 70.5 & 3.77 & 53.5 & 3.42 \\
      Vicuna-7B & D2 & 71.2 & 3.80 & 53.7 & 3.45 \\ \midrule
      Vicuna-13B & D1 & 71.1 & 3.79 & 54.2 & 3.45 \\   
      Vicuna-13B & D2 & \cellblue{\textbf{71.8}} & \cellblue{\textbf{3.80}} & \cellblue{\textbf{54.5}} & \cellblue{\textbf{3.46}} \\  
      \bottomrule
    \end{tabular}}
    \end{center}        
    \end{minipage}
    \hspace{2mm}
    \begin{minipage}{0.6\textwidth}
    \begin{center}
    \caption{Impact of different versions of GPT-3.5 Turbo on MSVD-QA and ActivityNet-QA. N/A signifies a version never set as GPT-3.5-Turbo default.}
    \label{tab:gpt_version}   
    \vspace{1mm}
    \setlength{\tabcolsep}{1pt}
    \scalebox{0.88}{
    \begin{tabular}{lccccc}
    \toprule
    \multirow{2}{*}{GPT-3.5-Turbo} & \multirow{2}{*}{Version Period} & \multicolumn{2}{c}{\bf MSVD-QA} & \multicolumn{2}{c}{\bf ANet-QA} \\
     & &  Acc & Score &  Acc & Score \\ \midrule
    \texttt{GPT-3.5-Turbo-0301} & 23/03/01-23/06/26 & \cellblue{82.4} & 3.99 & \cellblue{59.1} & \cellblue{3.59}  \\
    \texttt{GPT-3.5-Turbo-0613} & 23/06/26-24/02/15 & 71.8 & 3.80 & 54.5 & 3.46 \\
    \texttt{GPT-3.5-Turbo-1106} & N/A & 74.4 & 3.95 & 54.4 & 3.52  \\
    \texttt{GPT-3.5-Turbo-0125} & 24/02/16-Present & 74.4 & \cellblue{4.06} & 51.6 & 3.54 \\
    \bottomrule
    \end{tabular}}         
    \end{center}
    \end{minipage}
\end{table}

\textbf{Further Discussion on Aggregation Methods.}  In addition to the two aforementioned methods, we offer more aggregation methods in Table~\ref{tab:more_agg} for readers' reference. First, we find that aggregating the visual tokens after projection leads to better performance, likely because the features are already aligned with the LLM at this stage. Regarding sparse aggregation, besides temporal mean pooling, we also tried pooling in the spatial dimension, but the results were significantly poor. This underscores the importance of maintaining the original patch tokens for each frame. Consequently, combining temporal and spatial aggregation did not yield better results.
For dense aggregation, its effectiveness is confirmed. Given the LLM's token input limit, we considered compressing the number of patch tokens per frame to incorporate more frame information. We used max pooling with a stride of 2 to reduce the number of patch tokens per frame while ensuring the pooled tokens originated from the original tokens. This approach led to further improvement.
In conclusion, the key to extending existing offline image MLLMs to the video domain lies in incorporating more frame information without altering the features of the original image-level patch tokens.

\textbf{Is Existing Video Instruction Tuning Truly Effective?} We revisit the well-known VideoChatGPT~\cite{videochatgpt}, which firstly established a quantitative benchmark for video MLLMs, introduced the VideoInstruct100K dataset for video instruction tuning, and utilized it to fine-tune the projection of LLaVA-v1~\cite{llava}, setting a widely adopted baseline. Following VideoChatGPT, we also employ temporal mean pooling to aggregate tokens from multiple frames and use VideoInstruct100K to train LLaVA-1.5~\cite{llava1.5}. However, as indicated in Table~\ref{tab:off_vs_on}, the performance after training the projection is subpar compared to directly using the original LLaVA-1.5. Moreover, when updating both the projection and LLM simultaneously, the performance further declines.
The VideoChatGPT paper only presents results after training and lacks a comparison with its initialized parameters (\ie, LLaVA-v1~\cite{llava}), so we further corroborate these additional results in Table~\ref{tab:videochatgpt}. Strikingly, using the trained parameters provided by VideoChatGPT results in inferior performance compared to using its initialized parameters directly (\ie, not training at all). This finding is incredibly surprising.
This suggests that VideoChatGPT may not have learned beyond the capabilities of the original LLaVA. 
Therefore, we also encourage video MLLM methods that employ video instruction tuning with existing image MLLMs (\eg, LLaVA series~\cite{llava,llava1.5}, MiniGPT4~\cite{minigpt4}) to reevaluate and verify whether the learned model indeed exceeds the original image MLLM.

\textbf{Scaling Results on LLMs.} LLaVA-1.5~\cite{llava1.5} offers two models with different sizes of LLMs: Vicuna-7B~\cite{vicuna} and Vicuna-13B. As shown in Table~\ref{tab:llm}, when the parameter size of the LLM increases from 7B to 13B, FreeVA achieves further performance improvement on both MSVD and ActivityNet datasets. Additionally, it's also observable that \texttt{D2} consistently exhibits slight enhancements over \texttt{D2}.

\textbf{The Significant Impact of GPT-3.5 Versions on Evaluation Results.} 
Existing studies have overlooked this crucial point, as they all directly follow the evaluation code provided by VideoChatGPT, which utilizes the ``gpt-3.5-turbo'' API for evaluation. 
However, as shown in Table~\ref{tab:gpt_version}, the specific version pointed to by "gpt-3.5-turbo" is subject to updates by OpenAI over time. Up to now, three different versions have been the default ``gpt-3.5-turbo'', and these variations in API versions significantly impact the results. Therefore, if not specified, the same model may yield vastly different results when tested at different times. For instance, the performance evaluated using \texttt{GPT-3.5-Turbo-0301} is significantly higher than other versions. We strongly urge that all video MLLM methods provide specific GPT-3.5 version to ensure fair comparisons with other methods.

\subsection{Main Results}

\begin{table*}[!t]
 \centering
  \caption{Comparison with SOTA methods on zero-shot video QA datasets. We recognize that the default version of GPT-3.5-Turbo periodically, significantly impacting performance evaluation. Thus, we also report the possible GPT-3.5 versions used for evaluation. ``MAR'' denotes \texttt{GPT-3.5-Turbo-0301} version, ``JUN'' denotes  \texttt{GPT-3.5-Turbo-0613} version, and ``JAN'' denotes the latest \texttt{GPT-3.5-Turbo-0125} version. Unless specific, we use $\whred{^*}$ to signify the likely version based on the paper's timing. 
 $^\dag$ denotes the results reported in \cite{videochatgpt}. \emph{To ensure fair comparison and support the growth of the community, we hope future work will specify the GPT-3.5 version.}
  }
 \label{tab:main_qa}
 \scalebox{0.87}{
 \setlength{\tabcolsep}{3pt}
\begin{tabular}{llccccccccc}
  \toprule
  \multirow{2}{*}{ Method} & \multirow{2}{*}{\shortstack{LLM \\Size}} & \multirow{2}{*}{\shortstack{GPT-3.5 \\ Version}} & \multirow{2}{*}{\shortstack{Video-Text \\ Train Data}} & \multirow{2}{*}{\shortstack{Train \\ Free}} & \multicolumn{2}{c}{\bf MSVD-QA} & \multicolumn{2}{c}{\bf MSRVTT-QA}  & \multicolumn{2}{c}{\bf ANet-QA}  \\
  &  & &  & & Acc & Score & Acc & Score & Acc & Score \\
  \midrule
  FrozenBiLM~\cite{frozenbilm}$^\dag$ & 7B & MAR$\whred{^*}$ & 10M & \textcolor{red}{\xmark} & 32.2 & --  & 16.8 & --  & 24.7 & --  \\
  Video-LLaMA~\cite{videollama}$^\dag$ & 7B &  MAR$\whred{^*}$ & 2M+11K & \textcolor{red}{\xmark} & 51.6 & 2.5  & 29.6 & 1.8  & 12.4 & 1.1 \\
  VideoChat~\cite{videochat}$^\dag$ & 7B & MAR$\whred{^*}$ & 10M+11K & \textcolor{red}{\xmark} & 56.3 & 2.8  & 45.0 & 2.5  & 26.5 & 2.2 \\
  Video-ChatGPT~\cite{videochatgpt}$^\dag$ & 7B & MAR$\whred{^*}$ & 100K & \textcolor{red}{\xmark} & 64.9 & 3.3  & 49.3 & 2.8  & 35.2 & 2.7 \\ 
  Valley~\cite{luo2023valley} & 7B & MAR$\whred{^*}$ & 702K+73K & \textcolor{red}{\xmark}  & 65.4 & 3.4  & 45.7 & 2.5  & 42.9 & 3.0 \\ 
  VaQuitA~\cite{wang2023vaquita} & 7B & MAR & 100K & \textcolor{red}{\xmark} &  74.6 & 3.7 & 68.6 & 3.3 & 48.8 & 3.3 \\ \hdashline
  Chat-UniVi~\cite{jin2023chatunivi} & 7B & JUN$\whred{^*}$ & 100K & \textcolor{red}{\xmark}  & 65.0 &  3.6  & 54.6 & 3.1  & 45.8 & 3.2 \\
  BT-Adapter~\cite{btadapter} & 7B & JUN$\whred{^*}$ &  100K & \textcolor{red}{\xmark}  & 67.5 &  3.7  & 57.0 & 3.2  & 45.7 & 3.2 \\
  LLaMA-VID~\cite{li2023llamavid} & 7B & JUN$\whred{^*}$ & 232K+107K & \textcolor{red}{\xmark} & 69.7 & 3.7  & 57.7 & 3.2  & 47.4 & 3.3 \\
  LLaMA-VID~\cite{li2023llamavid} & 13B & JUN$\whred{^*}$ & 232K+107K & \textcolor{red}{\xmark} & 70.0 & 3.7  & 58.9 & 3.3  & 47.5 & 3.3 \\
  Video-LLaVA~\cite{Video-LLaVA} & 7B & JUN$\whred{^*}$ & 702K+100K & \textcolor{red}{\xmark} & 70.7 & 3.9 & 59.2 & 3.5  & 45.3 & 3.3 \\
  MovieChat~\cite{moviechat}  & 7B & JAN$\whred{^*}$ & 2M+11K & \textcolor{red}{\xmark} & 75.2 & 3.8  & 52.7 & 2.6 & 45.7 & 3.4 \\
  \midrule
  \multirow{2}{*}{\textbf{FreeVA} \small{\emph{w/ LLaVA-1.5}}\cite{llava1.5}}  & 7B  & \multirow{2}{*}{MAR} & \multirow{2}{*}{N/A} & \multirow{2}{*}{\textcolor{green}{\cmark}} & 81.5  & 4.0 & 72.9 & 3.5 & 58.3  & 3.5 \\
  &  13B&  &  &  & 82.4 & 4.0 & 73.6 & 3.6 & 59.1 & 3.6  \\ \hdashline
  \multirow{2}{*}{\textbf{FreeVA} \small{\emph{w/ LLaVA-1.5}}\cite{llava1.5}} & 7B & \multirow{2}{*}{JUN} & \multirow{2}{*}{N/A} & \multirow{2}{*}{\textcolor{green}{\cmark}} & 71.2 & 3.8 & 58.5 & 3.2 & 53.7 & 3.5\\
   & 13B &  &  &  & 71.8 & 3.8 & 59.2 & 3.3 & 54.5 & 3.5 \\ \hdashline
  \multirow{2}{*}{\textbf{FreeVA} \small{\emph{w/ LLaVA-1.5}}\cite{llava1.5}}& 7B & \multirow{2}{*}{JAN} & \multirow{2}{*}{N/A} &  \multirow{2}{*}{\textcolor{green}{\cmark}}&  73.8 & 4.1 & 60.0 & 3.5 & 51.2 & 3.5 \\  
  & 13B &  &  &   &  74.4 & 4.1 & 61.1 & 3.6 & 51.6 & 3.5  \\ \midrule
   \textbf{FreeVA} \small{\emph{w/ InstructBLIP}}\cite{instructblip} & 13B & {JAN} & {N/A} &  \textcolor{green}{\cmark} & 57.8 & 3.5 & 34.6 & 2.7 & 23.0 & 2.4 \\
    \midrule
   \textbf{FreeVA} \small{\emph{w/ InternVL}}\cite{internvl-1.5} & 20B & JAN & N/A &  \textcolor{green}{\cmark} & 73.8 & 4.0 & 54.8 & 3.3 & 58.4 & 3.6 \\ \midrule
  \multirow{2}{*}{\textbf{FreeVA} \small{\emph{w/ DenseConnector}}\cite{dc}}& 13B & \multirow{2}{*}{JAN} & \multirow{2}{*}{N/A} &  \multirow{2}{*}{\textcolor{green}{\cmark}}& 75.1 & 4.1 & 60.8 & 3.5 & 52.6 & 3.5 \\  
  & 34B &  &  &   &  77.4 & 4.2 & 62.1 & 3.6 & 55.8 & 3.6  \\ 
  \bottomrule
\end{tabular}}
\end{table*}

\noindent\textbf{Zero-shot Video Question-Answering.}
As shown in \Cref{tab:main_qa}, we can observe significant performance variations when using different versions of GPT-3.5. 
To ensure fair comparisons with previous works, we present results obtained using various versions of GPT-3.5 for evaluation.
Using the MAR version, FreeVA with LLaVA-1.5 demonstrates absolute advantages over pioneering works such as VideoChatGPT~\cite{videochatgpt} and Valley~\cite{luo2023valley} across all three datasets. Compared to recent work like VaQuitA~\cite{wang2023vaquita} (which explicitly states the use of the MAR version of GPT-3.5),  FreeVA still demonstrates significant superiority.
For the JUN version, Video-LLaVA~\cite{Video-LLaVA} and LLaMA-VID~\cite{li2023llamavid}, despite incorporating additional video training data, show inferior performance compared to FreeVA. Specifically, Video-LLaVA involves 702k video-text pairs from Valley~\cite{luo2023valley} for pretraining and utilizes VideoInstruct100K for video instruction tuning, while LLaMA-VID involves 232k video-caption pairs sampled from the WebVid 2.5M dataset~\cite{webvid} and also utilizes VideoInstruct100K for video instruction tuning. However, FreeVA delivers outstanding performance without any video transfer training, particularly on the ActivityNet-QA dataset, which is surprising and thought-provoking.
MovieChat~\cite{moviechat} designs a complex short-term and long-term memory mechanism to extend Video-LLaMA~\cite{videollama} for long video understanding. In comparison, our simple FreeVA significantly outperforms MovieChat in both MSRVTT-QA and ActivityNet-QA.
Recently, the default version of GPT-3.5 switched to the 2024 JAN version in February.
Therefore, the results provided here serve as a reference for future works. 
Additionally, equipping FreeVA with the more advanced MLLM (\ie, InternVL~\cite{internvl-1.5}, Dense Connector~\cite{dc}) will result in significant further improvements in video understanding performance.

\begin{table*}[t]
 \centering
  \caption{Comparison with leading methods on the video-based generative performance benchmark~\cite{videochatgpt}, \ie, Correctness of Information (\textbf{CI}), Detail Orientation (\textbf{DO}), Contextual Understanding (\textbf{CU}), Temporal Understanding (\textbf{TU}), and Consistency (\textbf{CO}). $^\dag$ denotes the results reported in~\cite{videochatgpt}.}
 \label{tab:main_text}
 \scalebox{0.93}{
\begin{tabular}{lcccccccc}
  \toprule
  \multirow{2}{*}{Method} &  \multirow{2}{*}{\shortstack{LLM \\Size}} & \multirow{2}{*}{GPT-3.5} & \multirow{2}{*}{\shortstack{Train \\ Free}} & \multirow{2}{*}{\bf CI} & \multirow{2}{*}{\bf DO}  & \multirow{2}{*}{\bf CU} & \multirow{2}{*}{\bf TU} & \multirow{2}{*}{\bf CO} \\  \\ \midrule
  Video-LLaMA~\cite{videollama}$^\dag$ & 7B &  MAR$\whred{^*}$ & \textcolor{red}{\xmark} & 1.96 & 2.18  & 2.16 & 1.82  & 1.79 \\
  LLaMA-Adapter~\cite{llamaadapter}$^\dag$  & 7B & MAR$\whred{^*}$ & \textcolor{red}{\xmark} & 2.03 & 2.32  & 2.30 & 1.98 & 2.15 \\
  VideoChat~\cite{videochat}$^\dag$ &7B & MAR$\whred{^*}$ & \textcolor{red}{\xmark} & 2.23 & 2.50  & 2.53 & 1.94  & 2.24 \\
  Video-ChatGPT~\cite{videochatgpt}$^\dag$ & 7B &MAR$\whred{^*}$ & \textcolor{red}{\xmark} & 2.40 & 2.52  & 2.62 & 1.98 & 2.37 \\ 
  Valley~\cite{luo2023valley}  & 7B &MAR$\whred{^*}$  & \textcolor{red}{\xmark} & 2.43 & 2.13 & 2.86 & 2.04 & 2.45  \\
  BT-Adapter~\cite{btadapter}  & 7B &JUN$\whred{^*}$ & \textcolor{red}{\xmark}  & 2.68 &  2.69  & 3.27 & 2.34 & 2.46 \\
  LLaMA-VID~\cite{li2023llamavid} & 13B & JUN$\whred{^*}$ & \textcolor{red}{\xmark} & 3.07 & 3.05  & 3.60 & 2.58  & 2.63 \\
  MovieChat~\cite{moviechat}  & 7B & JAN$\whred{^*}$ & \textcolor{red}{\xmark} & 2.76 & 2.93  & 3.01 & 2.24 & 2.42 \\
  \midrule
  \multirow{3}{*}{\textbf{FreeVA} \small{\emph{w/ LLaVA-1.5}}\cite{llava1.5}}  & 13B &  MAR & \textcolor{green}{\cmark} & 2.88 & 2.52 & 3.25 & 2.32 & 3.07 \\
  & 13B &JUN & \textcolor{green}{\cmark}  & 2.90 & 2.52 & 3.26 & 2.32 & 3.07 \\
  & 13B &JAN & \textcolor{green}{\cmark} & 2.88  & 2.52 & 3.25 & 2.34 & 3.05 \\ \midrule
\textbf{FreeVA} \small{\emph{w/ InstructBLIP}}\cite{instructblip} & 13B & JAN & \textcolor{green}{\cmark} & 2.78 & 2.43 & 3.10 & 2.30 & 2.82 \\ \midrule
\textbf{FreeVA} \small{\emph{w/ InternVL}}\cite{internvl-1.5} & 20B & JAN & \textcolor{green}{\cmark} & 3.25 & 2.50 & 3.44 & 2.52 & 3.29 \\ \midrule
\multirow{2}{*}{\textbf{FreeVA} \small{\emph{w/ DenseConnector}}\cite{dc}} & 13B & JAN & \textcolor{green}{\cmark} & 2.85 & 2.53 & 3.23 & 2.29 & 2.96 \\
& 34B & JAN & \textcolor{green}{\cmark} & 3.00 & 2.53 & 3.25 & 2.65 & 2.92 \\
  \bottomrule
  \end{tabular}
  }
\end{table*}

\begin{figure}[t!]
    \centering
    \includegraphics[width=0.97\textwidth]{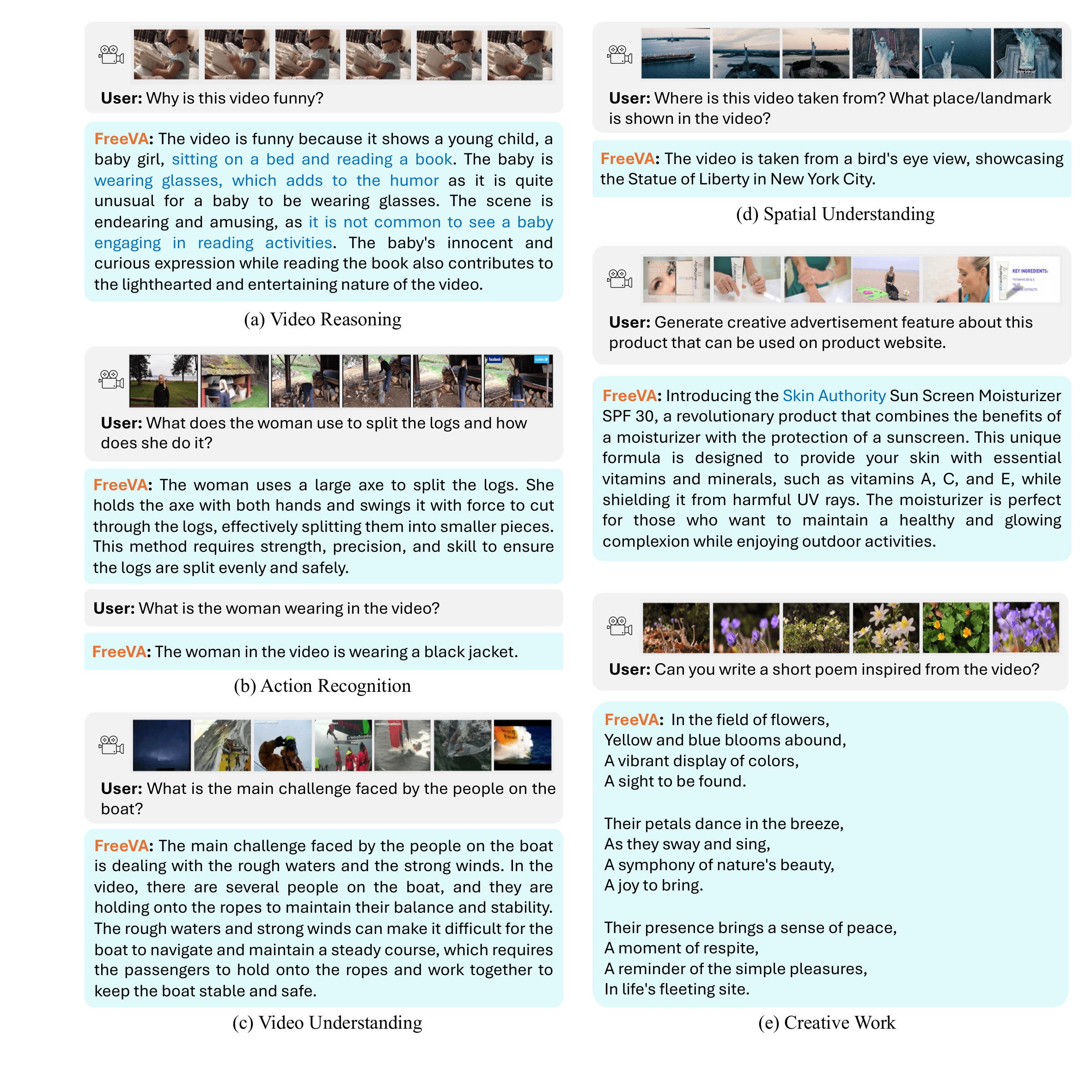}
    \caption{Examples of FreeVA's capabilities in multimodal video understanding, illustrating its proficiency in generating responses based on provided instructions. Please zoom in for best view.}
    \label{fig:vis}
\end{figure}

\noindent\textbf{Video-Based Text Generation Performance.}
As shown in Table~\ref{tab:main_text}, we can observe that the results on this benchmark remain relatively stable across different versions of GPT-3.5. As a method that requires no training, FreeVA continues to demonstrate promising performance, reaffirming the potential of directly applying image-based MLLM to video understanding.

\textbf{Qualitative Results.}
In Figure~\ref{fig:vis}, we showcase the qualitative results of FreeVA across various examples of video-based QA and conversational examples. Notably, even without any video instruction tuning, FreeVA still possesses the capability to understand videos. This includes video reasoning, action recognition, spatial and temporal understanding, and the ability to perform creative tasks such as composing poetry and creating advertisements. 

\section{Limitation and Discussion}
\label{sec:limitation}
FreeVA conducted a simple, intuitive, and effective study, demonstrating that by concatenating multiple visual tokens generated by MLLMs without any training. However, there are still areas worth further exploration:
1) Advanced Parameter-Free Strategies: Designing more sophisticated parameter-free strategies to compress single-frame information could allow for the inclusion of more frames within the token limit accepted by LLMs, enhancing video understanding performance.
2) Future Video Instruction Tuning: With larger-scale and higher-quality video instruction tuning data, models surpassing image-based MLLM knowledge for video tasks will likely emerge. Thus, FreeVA does not negate the importance of video tuning but aims to provide a simple strategy for offline MLLMs to serve as training-free video assistants. Additionally, incorporating the dense aggregation method proposed by FreeVA into the training process could potentially improve performance.
3) Expanding with More Advanced MLLMs: Currently, this study extends FreeVA using only a limited set of image MLLMs. With the rapid development of MLLMs, FreeVA should also benefit from future, more advanced image MLLMs for video understanding.

\section{Conclusion}
This paper introduces FreeVA, a study dedicated to exploring the direct extension of existing image-based MLLMs for video understanding without any additional training. FreeVA reveals some surprising phenomena overlooked by past research: leveraging only well-trained offline MLLMs without additional video fine-tuning excels in zero-shot video question-answering and even outperforms leading video models. Furthermore, using existing video instruction tuning data to tune MLLMs may not necessarily lead to improvements. Additionally, default version changes in GPT-3.5 can have a significant impact on performance evaluation in widely-used benchmarks. We hope that this work will provide valuable insights and experiences for future research and serve as a plug-and-play, simple yet effective baseline to encourage the direct evaluation of existing MLLMs on video tasks. 
We do not foresee negative social impact from the proposed work.

\newpage
\bibliographystyle{unsrt}
\bibliography{reference}

\end{document}